\documentclass[12pt]{article}

\usepackage{newtxtext,newtxmath}
\usepackage{graphicx}
\usepackage[letterpaper,margin=1in]{geometry}

\renewenvironment{abstract}
	{\quotation}
	{\endquotation}

\date{}

\makeatletter
\renewcommand{\fnum@figure}{\textbf{Figure \thefigure}}
\renewcommand{\fnum@table}{\textbf{Table \thetable}}
\makeatother

\usepackage{scicite}

\usepackage{url}

\def\scititle{
    Learning to infer and manipulate through distributed whole-arm interaction in a soft robot
}
\title{\bfseries \boldmath \scititle}

\author{
	Chuhan~Zhang$^{1}$,
	Ebrahim~Shahabi$^{1}$,
    Kseniia~Khomenko$^{1}$,\and
    Wei~Pan$^{2}$,
    Cosimo~Della~Santina$^{1}$\and
	\small$^{1}$Cognitive Robotics Department, Faculty of Mechanical Engineering, \and 
    \small Delft University of Technology, Delft 2628 CD, The Netherlands.\and
	\small$^{2}$School of Engineering, Newcastle University, UK.\and
	\small$^\ast$Corresponding author. Email: C.Zhang-8@tudelft.nl
}

\begin{document} 

% Insert the title and author list
\maketitle

% Abstract, in bold
% There are strict length limits, and not all formats have abstracts.
% Consult the journal instructions to authors for details.
% Do not cite any references in the abstract.
\begin{abstract} \bfseries \boldmath
In animals such as elephants and octopuses, acquiring non-visual information about an object and physically engaging with it are inseparable processes mediated by rich, large-area interactions between compliant appendages and the environment. Soft robots provide a natural platform for translating this principle into engineered systems. Yet current robotic intelligence makes limited use of physical interaction, treating it primarily as a disturbance to be rejected or, at best, as a means of compensating for object misalignment.

Here, we introduce a physical intelligence framework in which distributed compliant interactions jointly reveal task-relevant information and organize manipulation behavior. This results in an intrinsically partially observable problem: key task-relevant information is never measured directly, but must instead be inferred from the history of physical interactions. We propose a reinforcement-learning architecture that addresses this challenge by learning a memory-based control policy end-to-end. The key innovations making this possible are 
% [secret technical sauce here].
% a single learning strategy in which the policy learns to explore so as to generate physical interactions, while taszk reward shapes manipulation based on the resulting interaction history.
(i) a pretrained exploration policy that provides a reference for broad workspace exploration, (ii) joint optimization that integrates exploration and grasping objectives within a single recurrent policy, and (iii) a two-stage sim-to-real adaptation including observation mapping and policy fine-tuning.

We demonstrate this principle through blind whole-arm grasping with a hybrid rigid-soft robotic arm that we equip with IMUs embedded directly within its compliant structure, providing its only source of proprioceptive sensing. The learned policy successfully identifies and grasps various objects by autonomously coordinating workspace exploration, object encounter and localization, inference of grasp-relevant properties, and stable whole-arm wrapping.

\end{abstract}

\section*{INTRODUCTION}

\begin{figure}
    \centering
    \includegraphics[width=\linewidth]{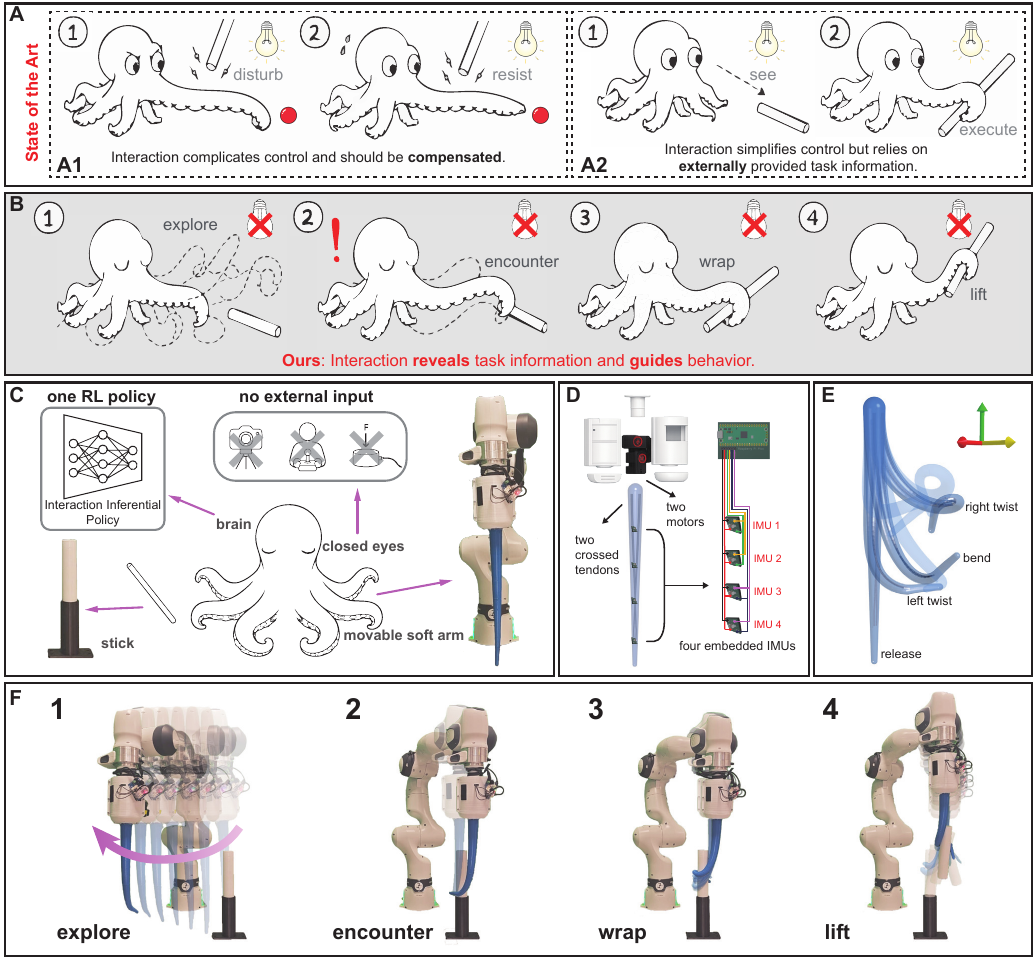}
    \vspace{-10mm}
    % \captionsetup{font=footnotesize,skip=2pt}
    \caption{
    \textbf{Overview of proposed Interaction Inferential Learning (IIL) framework for blind grasping with a proprioceptive soft robotic arm.} (\textbf{A}) Existing control paradigms for handling physical interaction. (\textbf{A1}) Interaction is treated as a disturbance to be compensated. (\textbf{A2}) Interaction facilitates task execution, whereas task-relevant information is supplied externally. (\textbf{B}) In IIL, interaction progressively reveals task information and organizes manipulation from exploration to grasping, without any external sensory input. (\textbf{C}) System overview. A single IIP controls the soft robotic arm to blind grasp a stick using only proprioceptive observations. (\textbf{D}) Hardware implementation with two crossed tendons and four embedded IMUs. (\textbf{E}) Representative deformation generated by tendon actuation. (\textbf{F}) Representative IIL manipulation trail consisting of exploration, contact, wrapping, and object lifting.
}
    \label{fig:abstract}
\end{figure}

% Animals + Existing robotics generally do not operate this way.
Muscular hydrostats, such as elephant trunks and octopus arms, enable dexterous manipulation in cluttered and unpredictable environments. Elephants retrieve hidden objects through a single access hole using only their trunks \cite{preti2023sensorized}; octopuses locate and grasp food with the arm alone when vision is severely limited \cite{buresch2022contact}. Unlike rigid skeletal systems, these compliant bodies continuously deform through contact, generating both mechanical constraints and distributed sensory feedback. Interaction with the environment is central to these tasks as it simultaneously shapes body configuration, reveals task-relevant information, and guides following actions. Therefore, intelligent behavior emerges from the coupling between body, sensing, and the environment, without relying on explicit external descriptions of the task. This view, however, differs fundamentally from most robotic manipulation systems, in which interaction is usually treated either as something to avoid or a disturbance to be compensated.

Soft robotic arms, which are robotic manipulators constructed from compliant materials rather than rigid links, provide a promising platform for realizing this biological principle, as their inherent compliance enables rich physical interaction. Recent advances have integrated sensing, actuation, and closed-loop control into increasingly capable soft robotic systems \cite{rus2015design, whitesides2018soft, trivedi2008soft, laschi2016soft, bang2024bioinspired, wang2024sensing, guan2023trimmed}. Their compliant architectures have also demonstrated robust grasping across diverse object geometries \cite{mazzolai2019octopus, wang2025spirobs, huang2025grasping, del2026peripheral, liu2026underwater}. 
However, realizing the full potential of soft bodies requires not only compliant hardware but also control strategies that exploit their embodied intelligence during physical interactions.

Existing control methods can be broadly divided into two perspectives. One treats environmental interaction as a disturbance, uncertainty, or source of model mismatch that must ultimately be compensated in order to maintain accurate control (Fig.~\ref{fig:abstract}A1). Such compensation may rely on explicit measurement of interaction through force or pose sensing \cite{della2023model,della2020model,fischer2023dynamic,franco2021energy,li2024disturbance,centurelli2022closed}, estimation through learned or adaptive dynamics models \cite{chen2024data,liu2025data, ho2018localized,wang2025robust,bruder2025koopman,habich2025generalizable,tang2026learning}, or adaptive and reinforcement learning policies \cite{tang2026general,satheeshbabu2019open,nazeer2024rl,jitosho2023reinforcement,gan2022reinforcement,thuruthel2018model,liu2023underwater,hou2026quantitative,zhang2026reinforcement}. Although these methods differ substantially in implementation, they share the same underlying goal: to minimize the effect of interaction on the desired control behavior. In this control paradigm, removing interaction from the system generally simplifies the control problem.

More recent advanced work instead exploits compliance and physical interaction as resources that simplify manipulation rather than sources of uncertainty (Fig.~\ref{fig:abstract}A2).
Alessi et al. \cite{alessi2024pushing} trained an RL policy to perform tip pushing tasks using externally specified target and tip positions, with the objective of minimizing their distance while maximizing contact force.
Jiang et al. \cite{jiang2021hierarchical} controlled the soft arm tip toward predefined motion targets while exploiting compliance to accommodate environmental constraints within a human-designed task strategy.
% Jiang et al. \cite{jiang2021hierarchical} used environmental contact to shape externally specified behaviors according to mechanical task constraints, for example, by opening a drawer using a provided handle location and a predefined backward movement.
Given the robot and object states, Johnson et al. \cite{johnson2025zero} trained an RL policy to coordinate two soft arms and the torso for box lifting, with a hand-designed motion primitive providing behavioral guidance during training. Montero et al. \cite{montero2024mastering} generalized human-demonstrated manipulation behaviors to visually specified task configurations through imitation learning.
% Johnson et al. \cite{johnson2025zero} used contact along two soft continuum arms and the torso to grasp and lift boxes of different sizes and masses, with the policy guided by a hand-designed motion primitive and externally predefined object states.
% Montero et al. \cite{montero2024mastering} exploited interaction-induced deformation to accommodate execution errors when generalizing manipulation behaviors from a human demonstration to visually prespecified task configurations. 
Teleoperation approaches similarly exploit compliance to improve manipulation robustness under human-in-the-loop control \cite{xie2023octopus,huang2026physical}. These studies establish interaction as a mechanical foundation that facilitates manipulation, rather than only a disturbance that limits performance.

However, despite exploiting compliance and environmental interaction to facilitate manipulation, existing control paradigms remain limited in addressing manipulation tasks in which the information required for action selection should be acquired through physical interaction, such as contact-guided object localization, exploratory manipulation, or manipulation under visual occlusion. The existing methods are formulated under the assumption that the information necessary for decision-making is already available before interaction begins, and is therefore explicitly provided in the observation space, such as target object states, task progress variables, or human-generated commands. In this way, physical interaction contributes to task execution but not to generating the information required for subsequent decisions. Learning, therefore, primarily serves to optimize behavior under a predefined informational structure rather than to establish that structure through interaction.

This limitation points to a broader scientific question in soft robotics: \textit{can physical interaction itself become a primary source of task-relevant information for intelligent manipulation?} Recent perspectives identify understanding how adaptive behavior emerges from the coupling between compliant bodies, sensing, and the environment as a central open challenge for the field \cite{laschi2026soft}. In particular, information self-structuring through sensorimotor coordination, whereby interaction with the environment shapes the information available to the controller, remains underexplored in the context of soft robotic control.

In this work, we introduce interaction inferential learning (IIL), a control paradigm in which physical interaction simultaneously reveals task-relevant information and organizes manipulation behavior (Fig. 1B). Rather than compensating for interaction or relying on externally prescribed task variables, IIL continuously infers what should be perceived, how the task should proceed, and when behavior should transition directly from ongoing physical interaction. 
To investigate this interaction-driven control paradigm, we use blind grasping as an experimental model because it isolates the informational role of physical interaction (Fig.~\ref{fig:abstract}C). Unlike conventional grasping with externally available object state, blind grasping requires the controller to determine where an object is, how it should be grasped, and when grasp acquisition has been achieved only through physical interaction. Accordingly, no external information, such as object poses, task progress variables, or human-generated commands, is provided to the controller. Instead, all these task-relevant information must emerge from the continuous coupling between the compliant body, proprioceptive sensing, and the environment.

We implement IIL on a proprioceptive soft robotic platform that integrates a tendon-driven soft arm with distributed embedded sensing and a single interaction inferential policy (IIP)(Fig.~\ref{fig:abstract}D,E). Operating only on proprioceptive observations, the policy autonomously organizes exploration, encounter with the object, and whole-arm stable grasping without manually designed behavioral stages or external task inputs (Fig.~\ref{fig:abstract}F). Across diverse object configurations, the resulting system successfully localizes and grasps objects through interaction alone. Our results demonstrate that the continuous coupling between a compliant body, proprioceptive sensing, and the environment can transform physical interaction from a mechanical consequence of manipulation into the process through which task-relevant information is generated and intelligent behavior is organized.

\begin{table}
\centering
\caption{\textbf{Comparison of control paradigms for soft robotic manipulation.}Whole-body manipulation is defined as manipulation involving physical interaction between the object and the soft arm body. In this table, tasks classified as non-whole-body manipulation involve object interaction only at the arm tip or through an external gripper. The role of interaction is not classified for purely human-designed or human-operated task control, where the relationship between interaction and task behavior is explicitly specified by the human designer rather than determined by the autonomous controller.}
\label{tab:comparison}

\scriptsize
\setlength{\tabcolsep}{2pt}
\renewcommand{\arraystretch}{1.15}

\begin{tabular}{
@{}
p{0.04\linewidth}
p{0.22\linewidth}
p{0.07\linewidth}
p{0.21\linewidth}
p{0.19\linewidth}
p{0.22\linewidth}
@{}
}
\hline
Work
& Task
& Whole-body
& Task control
& External task information
& Role of interaction \\
\hline

\cite{huang2025grasping}
& Grasp
& Yes
& Human-designed rules
& Grasp strategy
& -- \\

\cite{del2026peripheral}
& Grasp
& Yes
& Human-designed rules
& Grasp strategy
& -- \\

\cite{xie2023octopus}
& Grasp
& No
& Human operator
& Human command
& -- \\

\hline

\cite{fischer2023dynamic}
& Tip position/force tracking
& No
& Preset model-based control
& Position/force reference
& Disturbance \\

\cite{tang2026learning}
& Tip position/force tracking
& No
& Preset adaptive control
& Position/force reference
& Disturbance \\

\cite{zhang2026reinforcement}
& Tip pose tracking
& No
& Preset RL control
& Tip pose trajectory
& Disturbance \\

\hline

\cite{alessi2024pushing}
& Tip pushing
& No
& RL policy
& Object pose and size
& Execution facilitation \\

\cite{jiang2021hierarchical}
& Constrained tip motion
& No
& Human-designed hierarchy
& Task strategy
& Execution facilitation \\

\cite{johnson2025zero}
& Grasp
& Yes
& Imitation-guided RL policy
& Object state
& Execution facilitation \\

\cite{montero2024mastering}
& Grasp
& Yes
& Imitation-learned policy
& Object position
& Execution facilitation \\

\hline

\textbf{Ours}
& \textbf{Blind grasp}
& \textbf{Yes}
& \textbf{RL policy}
& \textbf{None}
& \textbf{Information generation and behavior organization} \\

\hline
\end{tabular}
\end{table}

\section*{RESULTS}

To examine whether task-relevant information for blind grasping can emerge from the physical interaction, we developed a platform in which a tendon-driven, proprioceptive soft arm was mounted on a rigid robotic manipulator that provided horizontal translation within a predefined workspace. Although developing this proprioceptive arm required addressing the challenges of integrating sensors into a compliant soft structure, this hardware design is not the primary contribution of this work. Detailed hardware implementation is described in the Methods and Materials section and the Supplementary Materials.
The successful blind grasping task required the policy to generate combined base motion and arm deformation to locate, wrap, and stably grasp an object whose position was unknown to the controller. Throughout the experiments, the policy relied on proprioceptive observations from the embedded sensors, such that the task-relevant information had to emerge from ongoing physical interaction with the environment.
Fig.~\ref{fig:real_object} and Fig.~\ref{fig:success_overview}A summarize the results of the experiments. In the following subsections, we first present the system's blind whole-arm grasping performance, and then analyze how and why this behavior emerges in detail.

% [the design of the arm is novel to embed to IMUs inside the arm; we point out the mechanical design is quite novel, because it's hard to sensorize the soft arm, however it's not the main contribution of the paper, so we leave this part into methods and metarial]

% [first paragraph: summarize the context in the huge section, the details are added later]

\subsection*{Interaction inferential learning enables blind whole-arm grasping}

\begin{figure}
    \centering
    \includegraphics[width=\linewidth]{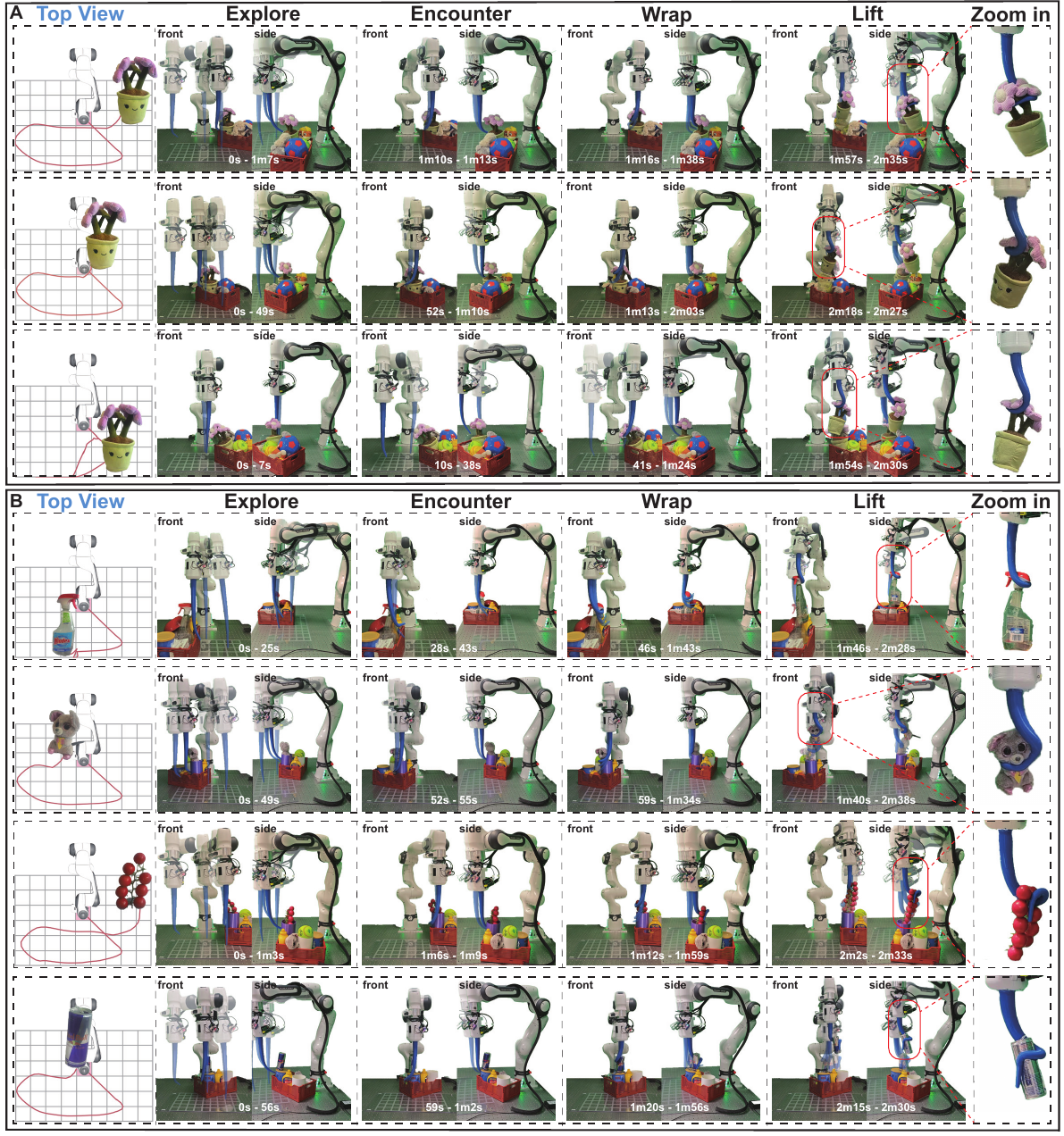}
    \caption{\textbf{Representative blind grasping trials performed by the proposed system.} (\textbf{A}) Successful grasping of the same toy flower placed at different locations within the workspace. (\textbf{B}) Successful grasping of four different daily life objects. For each trial, the top-view trajectory is shown together with representative snapshots of the exploration, encounter, wrapping, and lifting stages. The final column provides enlarged views of the grasped objects after lifting.}
    \label{fig:real_object}
\end{figure}

\begin{figure}
    \centering
    \includegraphics[width=\linewidth]{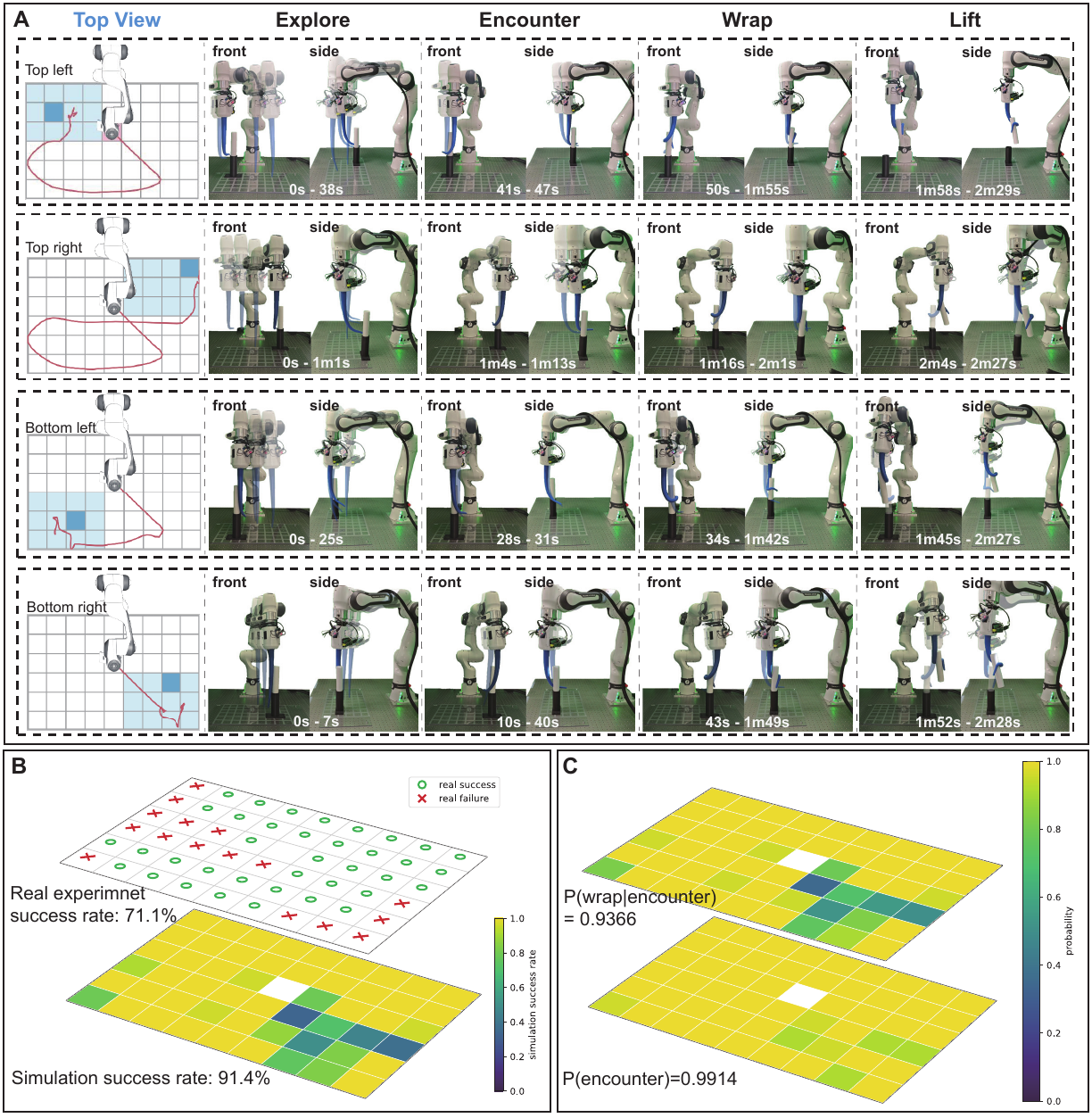}
   \caption{
    \textbf{Blind grasping performance across the reachable workspace.} (\textbf{A}) Representative successful grasping trials for cylindrical objects placed in four different workspace regions. For each trial, the top-view trajectory is shown. Representative snapshots of the exploration, encounter, wrapping, and lift stages are also provided. (\textbf{B}) The map uses color-coding to show the simulation's success rate for each object's location, while the real-world results are represented by green circles for success and red crosses for failure. (\textbf{C}) The map uses color-coding to show the spatial distribution of encounter probability and the conditional probability of successful wrapping given an encounter.
    }
    \label{fig:success_overview}
\end{figure}

Two experimental settings were used to evaluate blind whole-arm grasping. A cylindrical object was used for policy training and quantitative evaluation because its standardized geometry enabled consistent comparison across trials, whereas a set of irregular real-world objects was used for qualitative assessment of the system's grasping capability across diverse object geometries. In both settings, objects were placed at different locations throughout the reachable workspace.
At the beginning of every trial, both the robot base and the soft arm were reset to the same initial configuration, with the arm hanging freely under gravity. Each manipulation trial lasted 140 s, after which the robot executed a predefined lifting motion to evaluate grasp stability. A trial was considered successful only if the object remained securely wrapped throughout the lift.
Unless otherwise specified, all experiments and analyses reported below were performed using the same IIP trained on the cylindrical object.

Qualitative experiments evaluated blind grasping under two conditions: different object placements within the workspace and diverse unseen real-world objects (Fig.~\ref{fig:real_object}). Fig.~\ref{fig:real_object}A shows the same toy flower placed at three different workspace locations. Although each placement produced a different encounter time and initial contact configuration, the resulting interaction trajectories all converged to stable whole-arm wrapping and object lifting.
Fig.~\ref{fig:real_object}B further demonstrates successful blind grasping of four unseen real-world objects, including a detergent bottle, a toy mouse, an artificial tomato cluster, and a beverage can. These objects differed substantially in size, cross-sectional shape, surface structure, and wrapping affordances, illustrating qualitative generalization beyond the cylindrical training object.
The annotations \textit{Explore}, \textit{Encounter}, \textit{Wrap}, and \textit{Lift} describe the observed behavioral progression and do not correspond to predefined behavioral stages or transition signals within the policy. This annotation convention is adopted throughout the following figures.
 
We next systematically quantified blind grasping success across the reachable workspace using cylindrical objects (Fig.~\ref{fig:success_overview}). Representative successful trials from four workspace regions are shown in Fig.~\ref{fig:success_overview}A. Across the tested workspace, the policy achieved success rates of 91.4\% in simulation and 71.1\% on the real physical platform, with successful trials broadly distributed across object locations (Fig.~\ref{fig:success_overview}B). All real-world trials, together with their corresponding sensor signals and policy outputs, are provided in the Supplementary Materials.
To understand the remaining failures, we separately quantified successful object encounter and successful whole-arm wrapping following encounter. Initial object encounter occurred with a uniformly high probability across the workspace ($P(\mathrm{encounter})=0.9914$), whereas successful wrapping after contact was less reliable ($P(\mathrm{wrap}\mid\mathrm{encounter})=0.9366$) and exhibited greater spatial variation (Fig.~\ref{fig:success_overview}C). These results indicate that blind grasp failures arise primarily during stable whole-arm wrapping after encounter, rather than during initial object contact.

Qualitative demonstrations and quantitative evaluation together show that IIL enables blind whole-arm grasping using proprioceptive sensing alone, without externally provided task-relevant information. In the next subsection, we investigate how physical interaction organizes stable whole-arm grasping.

\subsection*{Interaction organizes different wrapping modes}

\begin{figure}
    \centering
    \includegraphics[width=\linewidth]{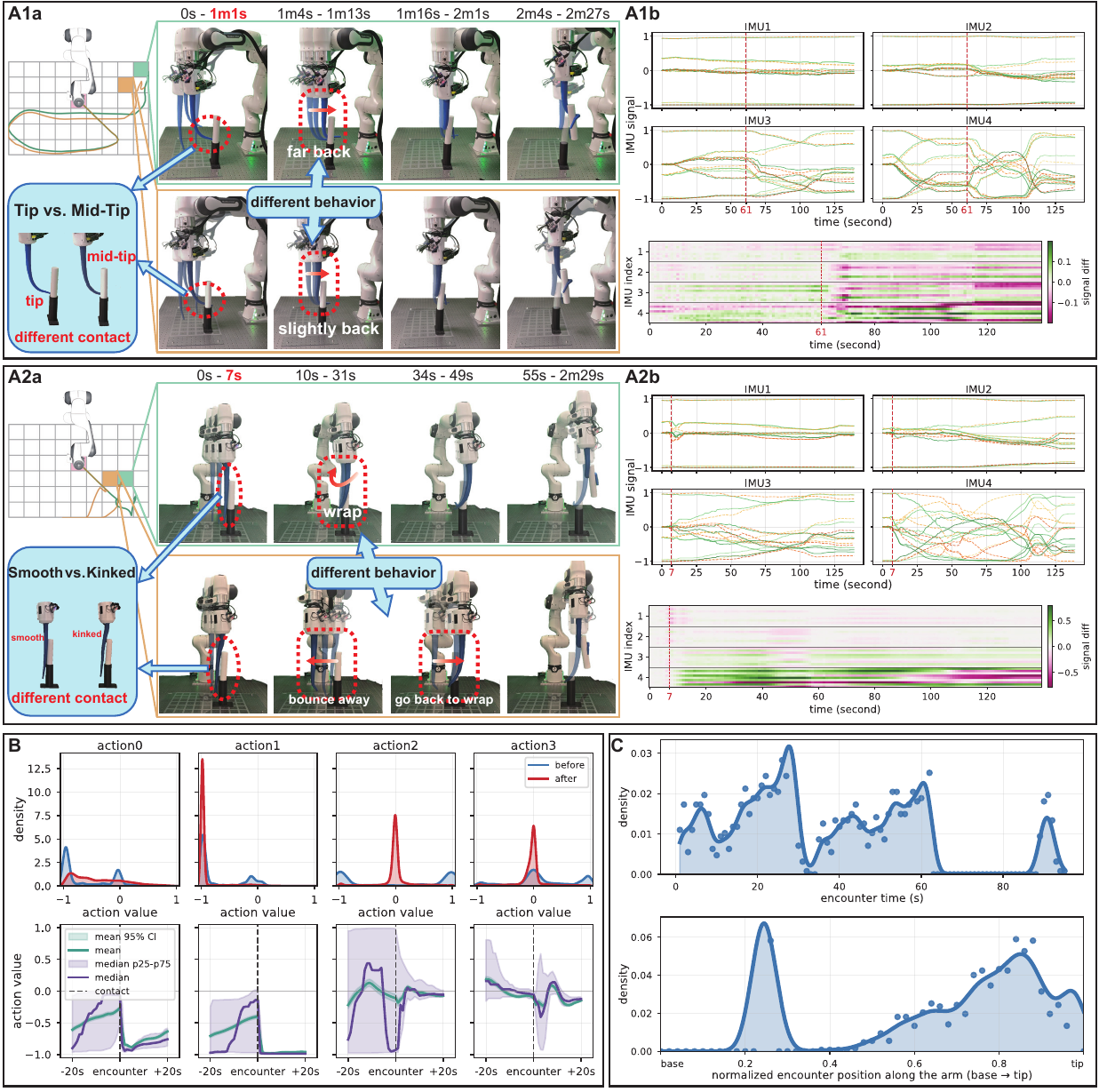}
    \vspace{-13mm}
    \caption{
    \textbf{Interaction-dependent organization of grasping behavior.} 
    (\textbf{A}) Representative trials show that different initial encounters reorganize subsequent wrapping behavior under the same IIP. Different encounter locations (\textbf{A1a}) and local contact geometries (\textbf{A2a}) generate distinct proprioceptive responses (\textbf{A1b}, \textbf{A2b}) and lead to different manipulation behaviors. Colors indicating locations in A1a and A2a correspond to the colors of IMU signals in A1b and A2b. 
    (\textbf{B}) Policy output statistics before and after object encounter. Kernel density estimates of the four action dimensions (top) and encounter-aligned action trajectories (bottom) show systematic changes following encounter. 
    (\textbf{C}) Distributions of first-encounter timing and arm location across trials, showing that encounters occurred at different times and at different locations along the soft arm.
}
    \label{fig:contact_behavior}
\end{figure}

\begin{figure}
    \centering
    \includegraphics[width=\linewidth]{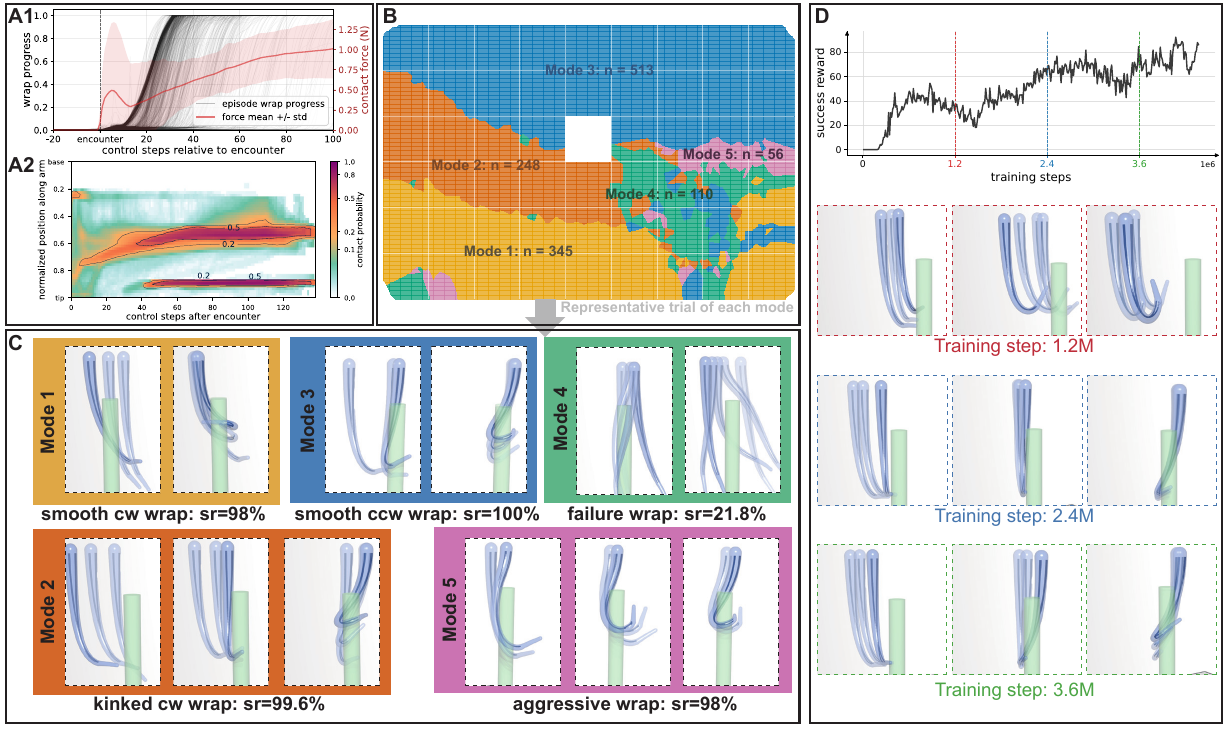}
    \caption{\textbf{Recurring interaction-driven wrapping modes emerging from IIL.} (\textbf{A}) Wrapping develops progressively after the first object encounter. Representative grasp-progress trajectories (\textbf{A1}) and contact-force evolution along the arm (\textbf{A2}) show sustained organization throughout the wrapping process. (\textbf{B}) Unsupervised K-means clustering of post-encounter interaction dynamics identifies recurring wrapping modes. We show the spatial distribution of 5 modes across the workspace.
    (\textbf{C}) Representative wrapping processes for each mode, illustrating different patterns of interaction-driven behavior. Modes 1 and 3 exhibit smooth wrapping, whereas Modes 2 and 5 involve arm bending or collision during wrapping but nevertheless result in successful grasps; Mode 4 represents unsuccessful wrapping. Clockwise (cw) and counterclockwise (ccw) indicate the wrapping direction, and sr denotes the grasp success rate of each mode.
    (\textbf{D}) Representative behaviors at different stages of training show how whole-arm wrapping gradually emerged during the IIL learning.}
    \label{fig:wrapping_modes}
\end{figure}

To understand how physical interaction organizes different wrapping behaviors, we analyzed how grasping evolved after different interaction events. Because interaction is a continuous process that cannot be naturally divided into discrete stages, we used object encounter as a representative interaction event for quantitative analysis. We then investigated how different encounter conditions reorganized subsequent grasping behavior and whether these behaviors converged into a small number of recurring wrapping modes across trials.

Representative trials showed that different encounters reorganized following grasping behavior, even though the same IIP governed all trials (Fig.~\ref{fig:contact_behavior}A). Encounters occurring at different arm locations (tip versus mid-tip in Fig.~\ref{fig:contact_behavior}A1a), or under different local contact geometries (smooth versus kinked in Fig.~\ref{fig:contact_behavior}A2a), produced distinct proprioceptive responses across the embedded IMUs (Fig.~\ref{fig:contact_behavior}A1b,A2b). The largest differences in sensor signals emerged immediately after encounter, indicating that different interactions generated different proprioceptive responses. Correspondingly, the policy produced different subsequent motions and wrapping behaviors despite having identical parameters. Quantitative analysis further showed that this interaction-dependent sensory signals were accompanied by systematic reorganization of the policy outputs (Fig.~\ref{fig:contact_behavior}B). Kernel density estimates revealed separate pre- and post-encounter action distributions across all four action dimensions (Fig.~\ref{fig:contact_behavior}B1), with a mean Jensen--Shannon divergence of 0.210 (95\% bootstrap CI, 0.205--0.216). Consistent with this observation, encounter-aligned action trajectories exhibited coordinated shifts across the four-dimensional action vector (Fig.~\ref{fig:contact_behavior}B2), with a median pre-to-post action displacement of 1.05 (IQR, 0.78--1.33). Importantly, object encounters occurred across a broad range of arm locations and episode times (Fig.~\ref{fig:contact_behavior}C), indicating that they were not associated with a fixed point in the control sequence. Despite this variability, the policy outputs remained consistently aligned to the moment of encounter rather than to absolute episode time. These results indicate that physical interaction continuously generated the sensory signals that reorganized subsequent grasping behavior rather than merely perturbing a fixed control trajectory.

Although different interactions gave rise to different grasping behaviors, these behaviors were not arbitrary but instead converged into a small number of recurring wrapping modes.
Whole-arm wrapping emerged progressively through continued interaction rather than instantaneously after the first encounter (Fig.~\ref{fig:wrapping_modes}A). Individual episodes varied in the onset of wrapping, yet exhibited a remarkably similar progression toward complete enclosure (Fig.~\ref{fig:wrapping_modes}A1). Likewise, initial contact occurred across a broad range of arm locations, but continued interaction progressively reorganized the contact-force distribution toward a similar whole-arm wrapping pattern (Fig.~\ref{fig:wrapping_modes}A2).
To characterize this convergence, we performed unsupervised clustering of post-encounter interaction dynamics based on whole-arm geometry and contact-force evolution, without using task-level variables such as grasp progress, task success, or object information. Details of the clustering procedure and feature construction are provided in the Supplementary Materials. This analysis identified five recurring wrapping modes (Fig.~\ref{fig:wrapping_modes}B). Four modes consistently produced successful whole-arm grasps despite following different interaction evolutions, whereas one mode, characterized by a strongly kinked initial contact, exhibited a substantially lower success rate (Fig.~\ref{fig:wrapping_modes}C). The spatial distribution of these modes across the workspace further suggests that different encounter conditions consistently gave rise to particular interaction modes rather than arbitrary behavioral variation. Moreover, these recurring wrapping modes emerged progressively during training instead of being explicitly programmed (Fig.~\ref{fig:wrapping_modes}D). Early policies frequently lost contact after encounter, intermediate policies completed unstable wrapping motions, and the final policy achieved smooth whole-arm wrapping. These results indicate that IIL progressively organized interaction-driven behaviors into a small set of stable and functional wrapping modes.

By showing that different interaction conditions systematically gave rise to different yet recurring grasping behaviors, these results establish physical interaction as the organizing principle underlying blind whole-arm wrapping. In the next subsection, we investigate why the IIL framework can learn this interaction-driven organization.

\subsection*{Interaction inferential learning establishes interaction-driven behavior}

\begin{figure}
    \centering
    \includegraphics[width=\linewidth]{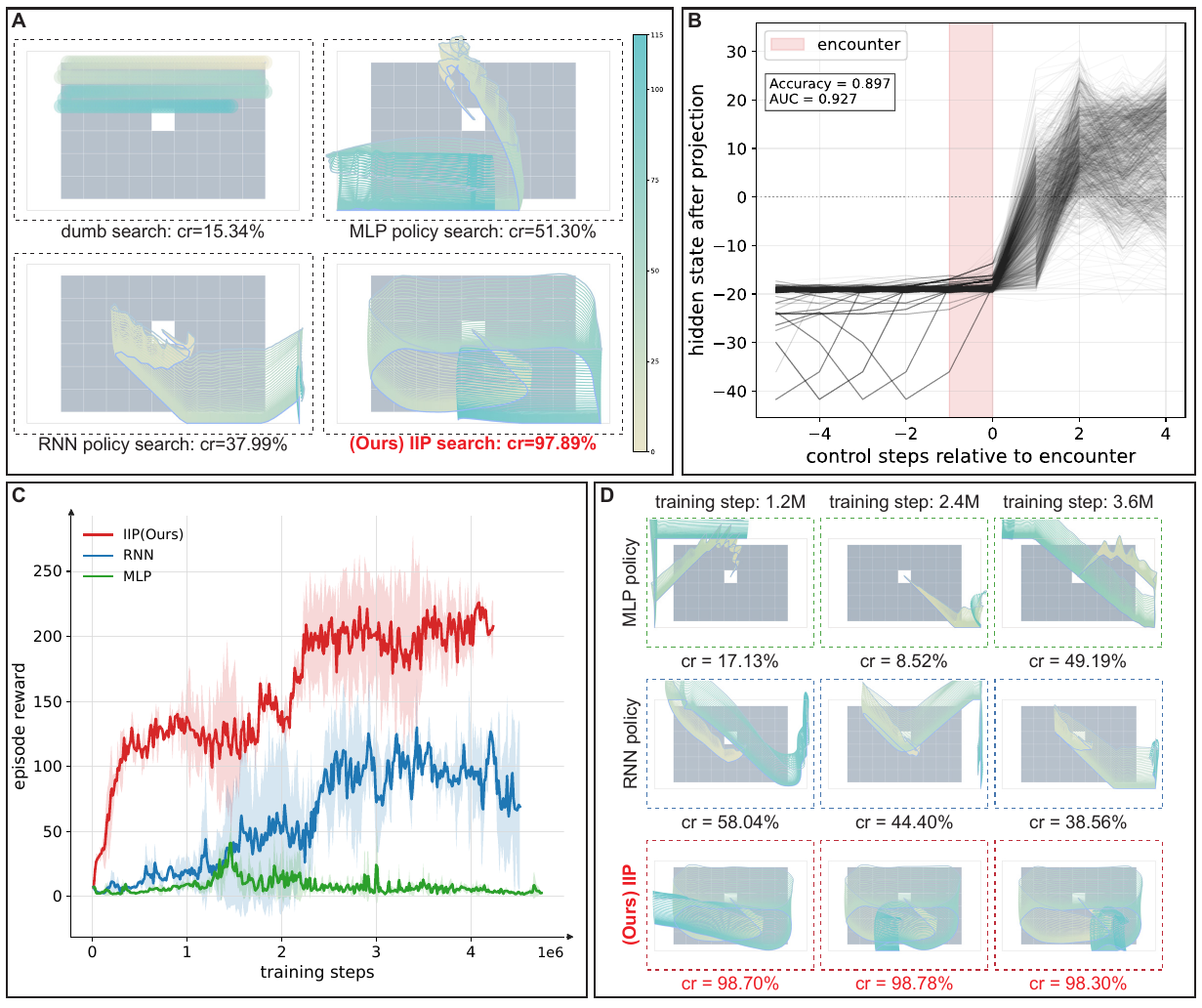}
    \vspace{-10mm}
    \caption{\textbf{Learning interaction-driven behavior through IIL.} (\textbf{A}) Workspace exploration trajectories of four policies: a predefined raster-scan policy, an MLP RL policy, a recurrent RL policy, and the IIP. The coverage rate (cr) quantifies the fraction of the workspace explored by each policy, with the IIP achieving the highest coverage. (\textbf{B}) Projection of recurrent latent states onto the discriminative direction of a linear classifier trained to distinguish pre-encounter and post-encounter representations. The projection shows a pronounced shift around the encounter, highlighting the corresponding change in latent representation. (\textbf{C}) Training reward curves of the MLP RL policy, recurrent RL policy, and IIP. The IIP achieves the highest reward during training. Shaded regions indicate the standard deviation across independent training runs. (\textbf{D}) Evolution of workspace exploration during training for the three learned policies, showing that only the IIP maintains high workspace coverage throughout training.}
    \label{fig:iil}
\end{figure}

In this subsection, we investigate why this interaction-driven organization emerges through IIL. Because IIL is a learning framework built upon a recurrent RL policy, we first examine the role of the recurrent architecture in representing physical interaction, and then investigate why the recurrent architecture alone is insufficient without the proposed learning framework. The detailed IIL framework is described in the Methods and Materials section.

We first examined whether the recurrent architecture underlying IIL is necessary to represent physical interaction. Compared with a feedforward multilayer perceptron (MLP) trained under the same environment setting, training budget, and network capacity as an ablation experiment, the IIP rapidly increased the reward throughout training, whereas the MLP showed little learning progress and maintained an almost flat reward curve (Fig.~\ref{fig:iil}C). This result indicates that without a recurrent architecture capable of integrating interaction history, the policy was unable to learn effective blind grasping behaviors. Consistent with this observation, only the recurrent IIP established broad exploration across the reachable workspace, whereas the MLP remained confined to a limited region throughout training (Fig.~\ref{fig:iil}A,D), suggesting that interaction-driven grasping requires information accumulated from previous interactions rather than only the current proprioceptive observation.
To determine whether physical interaction was consistently represented within the recurrent RL policy, we analyzed the latent state around the first encounter. Latent states before and after the first encounter were separated using a linear logistic classifier and projected onto the corresponding discriminative direction (Fig.~\ref{fig:iil}B). Details of the linear classification analysis are provided in the Supplementary Materials. Despite large variations in the timing and spatial location of encounters, the projected latent trajectories exhibited a remarkably consistent transition after encounter across trials. The classifier distinguished pre- and post-encounter latent states with an accuracy of 0.897 (AUC = 0.927), indicating that physical interaction induced a linearly decodable transition in the policy's latent representation. These results indicate that the recurrent architecture enables the policy to establish an internal representation of physical interaction, providing the representational basis on which IIL learns interaction-driven behavior.

However, the recurrent architecture alone was not sufficient to establish the interaction-driven organization. When the same recurrent RL policy was trained using conventional RL framework instead of IIL as another ablation experiment, exploration remained unstable and repeatedly collapsed to limited regions of the workspace (Fig.~\ref{fig:iil}A,D). As a result, the policy frequently failed to maintain repeated interaction with the object and therefore acquired only limited interaction experience throughout learning. In contrast, IIL progressively established broad and stable interaction coverage across the reachable workspace, allowing the recurrent policy to repeatedly experience diverse contact conditions and accumulate task-relevant information through physical interaction. Consistent with this progressively organized exploration, IIL achieved substantially higher episodic rewards than the recurrent ablation throughout training (Fig.~\ref{fig:iil}C). These results indicate that the interaction-driven organization does not emerge from recurrent memory alone. Instead, IIL provides the learning mechanism that progressively organizes recurrent interaction representations into stable blind whole-arm grasping behaviors.

These results indicate that within IIL, the recurrent architecture enables physical interaction to be represented in the policy's internal state, whereas the learning progressively organizes these representations into stable interaction-driven grasping behaviors.

\section*{DISCUSSION}

Physical interaction is typically regarded as a mechanical phenomenon in robotic manipulation. Depending on the control formulation, it is either compensated as a disturbance or exploited to simplify the execution of a task whose informational structure has already been externally specified. In contrast, our presented work demonstrates a different role of interaction. In blind grasping, where no external object states, task-progress variables, or human commands are available, successful manipulation requires the controller to acquire task-relevant information directly from ongoing contact. Our results show that, under this formulation, physical interaction is not merely part of the manipulation dynamics but becomes the process through which the information required for subsequent decisions is generated. This shifts the role of interaction from a mechanical consequence of control to an active component of intelligent behavior.

This formulation also changes how closed-loop control is organized. Conventional manipulation systems regulate the interaction between the robot and the environment under the assumption that sufficient task-relevant information is already available before execution begins. Under IIL, however, control not only determines the evolution of the physical system but also shapes the information available for subsequent decisions. Control commands, therefore, contribute simultaneously to changing the body-environment relationship and to generating the information that organizes future behavior. These observations suggest that in closed-loop control, physical regulation and information generation are inseparable aspects of the same process rather than sequential stages.

The applicability of this framework depends on both the task and the embodiment. At the task level, IIL is advantageous when physical interaction provides information that is unavailable before execution. When task-relevant information is already available at the beginning of control, conventional formulations remain appropriate because interaction no longer serves an inferential role. At the embodiment level, the effectiveness of this framework further depends on whether physical interaction produces sufficiently informative sensory responses. Soft robotic arms naturally satisfy this condition. Their compliant bodies undergo distributed deformation during contact, producing rich proprioceptive responses that encode both body configuration and environmental constraints. Rather than only serving as a mechanical advantage for safe or adaptive manipulation, compliance therefore also provides an informational substrate through which interaction can be interpreted and accumulated over time. From this perspective, embodiment contributes not only to how the robot acts on the environment but also to what the robot can infer from interacting with it.

Although demonstrated here in blind grasping, we expect the underlying principle to extend beyond this specific task. Many manipulation problems require information to emerge progressively through interaction rather than being available beforehand, including exploratory object localization, threading, or navigation within confined or uncertain environments. The common characteristic of these problems is not simply that they involve frequent contact, but that physical interaction constitutes an essential source of information required for subsequent decision-making. Extending IIL to these settings, together with richer representations of interaction-generated information and longer-horizon reasoning, may provide a broader framework for embodied manipulation in environments where explicit task-state information is unavailable.

\section*{MATERIALS AND METHODS}

We developed IIL to train a recurrent RL policy for blind grasping from proprioceptive interaction histories. The policy integrates successive proprioceptive observations through a recurrent hidden state and maps the resulting internal representation to tendon actuation and planar base motion. During training, IIL applies phase-structured joint training to this single policy: a reference-guided objective promotes workspace exploration before encounter, whereas an RL objective optimizes contact-dependent enclosure and grasp formation.
Although these training objectives are explicitly organized by physical task phases, this phase structure remains invisible to the policy itself. The resulting policy, termed the IIP, is therefore required to organize the entire manipulation process solely from the evolving proprioceptive observation history, rather than from explicit phase information.
Below, we first describe the IIP formulation and architecture, followed by the IIL training framework and its simulation-to-hardware implementation.

\subsection*{Interaction inferential policy}

The policy received proprioceptive observations from four sensing locations distributed along the soft arm. At each control step, the local orientation at each sensing location was represented by a $3\times3$ rotation matrix. Flattening the four rotation matrices produced a 36-dimensional observation vector. No additional task-related information, including object position, contact state, or interaction phase, was provided to the policy. As a result, the instantaneous observation at a single control step was insufficient to determine the manipulation state.
Instead, task-relevant information had to be inferred from the temporal evolution of proprioceptive observations generated during physical interaction. 

To accumulate this information over time, the policy was formulated as a recurrent network whose hidden state continuously integrated successive observations throughout the manipulation process. At control step $k$, its internal state was updated from the current observation $o_k$ and the preceding hidden state $h_{k-1}$ according to the recurrent mechanism
$
h_k=f_{\theta_h}(o_k,h_{k-1}),
$
and the action was sampled from a distribution conditioned on the updated hidden state,
$
a_k\sim\pi_{\theta_\pi}(\cdot\mid h_k).
$
Here, $f_{\theta_h}$ denotes the recurrent state-transition function and $\pi_{\theta_\pi}$ denotes the action distribution, with $\theta_h$ and $\theta_\pi$ jointly constituting the actor parameters $\theta$. The policy, therefore, represented a mapping from the observation history $o_{0:k}$, rather than from the instantaneous observation $o_k$, to the current action. The recurrent state was initialized to zero at the start of each episode and propagated through the entire manipulation sequence, allowing temporal information to influence following decisions.

The actor network consisted of a two-layer gated recurrent unit (GRU) with 256 hidden units per layer, followed by a multilayer perceptron with one hidden layer of 256 units.
The policy output consisted of four continuous control commands for the soft arm system. Two commands controlled the tendon actuation of the soft arm, and the remaining two controlled the planar translation of the arm base. The physical implementation of these control commands is described in the Hardware deployment section, and the detailed network architecture is provided in the Supplementary Materials.
The following section describes how this recurrent policy was optimized under the IIL framework.

\begin{figure}
    \centering
    \includegraphics[width=\linewidth]{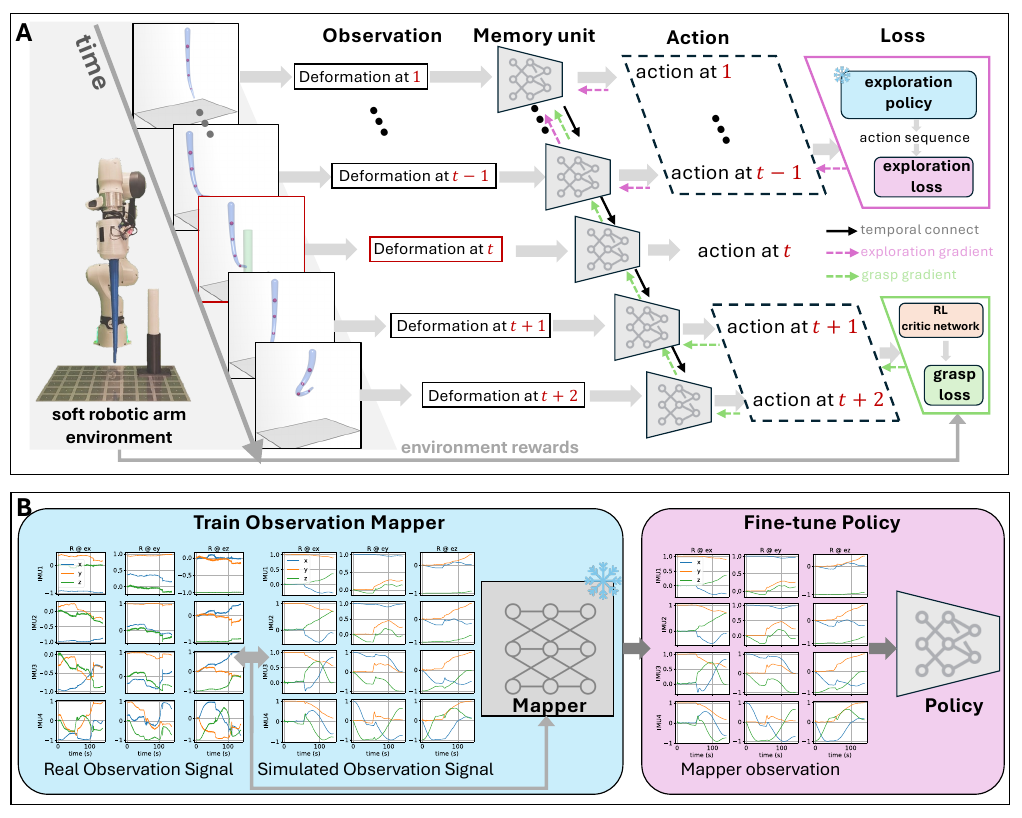}
    \caption{\textbf{Overview of the proposed IIL training framework.} (\textbf{A}) IIL framework. A single recurrent IIP is optimized using two local objectives associated with different interaction phases: a reference-guided exploration objective before object encounter and a reinforcement learning objective for contact-dependent grasping after encounter. Both objectives update the same recurrent policy throughout gradient backpropagation. (\textbf{B}) Sim-to-real transfer. A recurrent observation mapper is first trained to align real proprioceptive observations with the simulated observation space. The pretrained IIP is then fine-tuned using the mapped observations before deployment on the physical platform.}
    \label{fig:algo}
\end{figure}

\subsection*{Interaction inferential learning}

We proposed IIL to train the single recurrent policy for the whole process of blind grasping (Fig.~\ref{fig:algo}A). Successful blind grasping naturally depends on multiple interaction functions, each with a different role. Before a stable grasp can be formed, the arm should first explore the workspace and establish physical contact. Only after contact has been established can interaction-induced deformation be exploited to organize enclosure and grasp formation. Because these functions serve different purposes yet are all required for successful manipulation, IIL optimizes the same recurrent policy using different training objectives associated with different stages of the interaction process.

Each interaction function is associated with a local training objective that specifies the optimization target applied to the policy at the corresponding stage of interaction. Let $\ell_n(\theta;k)$ denote the $n$-th local actor objective at control step $k$. The overall actor objective is defined as
\begin{equation}
\mathcal{L}_{\mathrm{actor}}^{\mathrm{IIL}}(\theta)
=
\sum_{n=1}^{N}
\lambda_n
\mathbb{E}_{\tau\sim\mathcal{D}}
\left[
\sum_k
m_n(\phi_k;\tau)
\ell_n(\theta;k)
\right],
\label{eq:iil_actor}
\end{equation}
where $\lambda_n$ is the corresponding weighting coefficient, $\mathcal{D}$ denotes the replay distribution over interaction trajectories $\tau$, and $\phi_k$ denotes the physical interaction phase at control step $k$.
For the present blind grasping task, we set $N$ to 2 and defined two local objectives: a reference-guided exploration objective and a soft actor-critic (SAC) RL objective for contact-dependent grasping.
During optimization, each sampled transition was assigned to one of these local objectives according to both its physical interaction phase and the current stage of training. This assignment was represented by the binary mask $m_n(\phi_k;\tau)$. For the present task, all transitions after the first contact were optimized using the grasp objective. Before the first encounter, the proportion of transitions assigned to the exploration objective gradually decreased over training, allowing the shared policy to progressively shift its emphasis from exploration toward grasp optimization. The detailed assignment scheduler is described in the Supplementary Materials. The physical interaction phase $\phi_k$ was obtained from privileged simulation information only for objective assignment during training and was never included in the policy observation.

\subsubsection*{Exploration objective}

Before learning the grasp behavior, we first trained a reference exploration policy, denoted by $\pi_{\mathrm{exp}}$, using the same policy architecture as the IIP. The purpose of this policy was to generate broad and stable exploratory motion throughout the reachable workspace before contact was established. To encourage this behavior, the exploration reward combined workspace coverage with motion stability. The planar workspace was partitioned into smaller grid cells to quantify exploration coverage. The soft arm was rewarded for visiting previously unexplored cells, while excessive oscillation of the tip segment was penalized. Details of the exploration reward formulation and parameter settings are provided in the Supplementary Materials. This reference policy was optimized using SAC, after which its parameters were fixed.

During subsequent IIL optimization, this policy $\pi_{\mathrm{exp}}$ served as a behavioral reference for the reference-guided objective. At control step $k$, the corresponding local actor objective was defined as
\begin{equation}
\ell_{\mathrm{exp}}(\theta;k)
=
\frac{1}{N_a}
\left\|
\bar{\pi}_{\theta}(o_k,h_{k-1})
-
\bar{\pi}_{\mathrm{exp}}(o_k,h_{k-1}^{\mathrm{exp}})
\right\|_2^2,
\label{eq:reference_exploration_loss}
\end{equation}
where $N_a$ is the action dimension, $\bar{\pi}_{\theta}(o_k,h_{k-1})$ denotes the deterministic action produced by the trainable IIP, and $\bar{\pi}_{\mathrm{exp}}(o_k,h_{k-1}^{\mathrm{exp}})$ denotes the deterministic action produced by the fixed reference exploration policy. Here, $h_{k-1}$ and $h_{k-1}^{\mathrm{exp}}$ are the recurrent hidden states of the trainable and reference policies, respectively. As specified by the assignment mask $m_n(\phi_k;\tau)$ in Eq.~\ref{eq:iil_actor}, this objective was applied to pre-contact exploration control steps assigned to reference-guided training. Eligible pre-encounter exploration control steps assigned to the SAC objective were excluded from this loss.

\subsubsection*{Grasp objective}

The contact-dependent grasping objective was SAC loss. At control step $k$, the corresponding local actor objective was defined as
\begin{equation}
\ell_{\mathrm{grasp}}(\theta;k)
=
\mathbb{E}_{a_k \sim \pi_{\theta}(\cdot \mid o_k,h_{k-1})}
\left[
\alpha \log \pi_{\theta}(a_k \mid o_k,h_{k-1})
-
\min_{i=1,2}
Q_i(o_k,h_{k-1},a_k)
\right],
\label{eq:grasp_objective}
\end{equation}
where $\alpha$ is the SAC entropy-temperature coefficient, and
$Q_1$ and $Q_2$ are the two recurrent critic networks.
Each critic estimates the action value from the observation history
$o_{0:k}$ and the sampled action $a_k$. 
The critics were trained using standard SAC Bellman regression with rewards provided by the simulation environment. For grasp learning, the reward captured the desired physical characteristics of whole-arm grasping, encouraging distributed and stable object contact together with progressive enclosure while penalizing unstable or undesirable interaction patterns. These reward-related quantities were used only during training and were not included in the policy observation. Details of the reward formulation and parameter settings are provided in the Supplementary Materials. These quantities were used only to construct the training reward and were not included in the policy observation.

Through the critic network learning, the environmental reward was incorporated into the estimated action values. The actor was therefore guided toward actions associated with stable contact and enclosure through $Q_1$ and $Q_2$. This SAC learning objective was used for all post-contact actor transitions and for pre-contact transitions assigned by the mask $m$. The resulting gradients updated the same recurrent actor as the reference-guided exploration objective.

\subsection*{Deploy on the hardware}

To deploy the learned policy on the physical platform, three components were required: a proprioceptive soft robotic arm that provided the observation and actuation interface, a physics simulation used for IIL training, and a sim-to-real adaptation procedure that aligned real proprioceptive observations with the simulated observations. These components are described below.

\subsubsection*{Hardware experimental design}
The physical platform consisted of a tendon-driven soft robotic arm mounted on the end effector of a 7-degree-of-freedom Franka Emika Panda manipulator. The soft arm was actuated by two crossed tendons driven by DYNAMIXEL MX-28 servo motors. Four LSM6DSOX six-axis IMUs embedded along the arm provided proprioceptive sensing. Details of the soft robotic arm design and fabrication are provided in the Supplementary Materials.
Each IMU streamed linear acceleration and angular velocity measurements at approximately 38~Hz. The measurements from each IMU were processed independently by a Madgwick filter initialized from a stationary calibration procedure to estimate the sensor orientation. The resulting rotation matrices was concatenated into a 36-dimensional proprioceptive observation. This observation was subsequently processed by the trained observation mapper.
The recurrent IIP was evaluated at 1~Hz to generate a four-dimensional control action containing commands for the tendon actuators and the horizontal motion of the rigid manipulator. Tendon commands were converted into motor position targets through fixed scaling factors. Between consecutive policy updates, motor position targets were linearly interpolated and transmitted to the DYNAMIXEL position controllers at 5~Hz to ensure smooth tendon motion. The manipulator commands were supplied to a Cartesian PID controller operating at 100~Hz, while this controller tracked the commanded motion at 1~kHz.

\subsubsection*{Modeling soft robotic arm}

The soft robotic arm was modeled as a Cosserat rod, which represents the manipulator as a continuous deformable body capable of large bending, twisting, stretching, and shear deformation. The governing equations were solved using the PyElastica framework \cite{PyElastica}, which provides numerical integration of Cosserat rod dynamics with contact interactions. This formulation captures the continuum mechanics required for compliant manipulation while remaining computationally efficient for large-scale reinforcement learning.

The simulated system consisted of a tendon-driven soft arm interacting with a rigid cylindrical object. Tendon forces were distributed along the compliant arm to generate continuous whole-body deformation, while the arm base was allowed to translate within the Cartesian workspace in the simulator through a three-axis PID position controller.
The physics simulation was integrated using a fixed time step of $10^{-4}\mathrm{s}$, while the high-level control command from the IIP operated at $5\mathrm{Hz}$. Detailed simulation settings are provided in the Supplementary Materials.

\subsubsection*{Sim-to-real transfer}

The IIP was trained entirely in simulation using proprioceptive observations derived directly from the simulated soft-arm configuration. During physical deployment, however, the same observation representation could not be obtained directly. Instead, the proprioceptive observation had to be estimated from IMUs, making it vulnerable to sensor noise, mounting offsets, filtering dynamics, and hardware-specific deformation effects. In this way, to bridge this observation-domain gap, we adopted the two-stage sim-to-real adaptation procedure illustrated in Fig.~\ref{fig:algo}B. First, synchronized real-robot and simulated proprioceptive trajectories were collected under corresponding motions to train a recurrent observation mapper that transformed real observations into the simulated observation space. The pretrained IIP was then fine-tuned using the mapped observations while preserving the original control architecture. During deployment, both the observation mapper and the fine-tuned policy remained fixed, and no online parameter adaptation was performed.

%%%%%%%%%%%%%%%% REFERENCES %%%%%%%%%%%%%%%

\clearpage 
\bibliography{ref} % for a file named science_template.bib

@article{della2020model,
  title={Model-based dynamic feedback control of a planar soft robot: trajectory tracking and interaction with the environment},
  author={Della Santina, Cosimo and Katzschmann, Robert K and Bicchi, Antonio and Rus, Daniela},
  journal={The International Journal of Robotics Research},
  volume={39},
  number={4},
  pages={490--513},
  year={2020},
  publisher={Sage Publications Sage UK: London, England}
}

@article{jiang2021hierarchical,
  title={Hierarchical control of soft manipulators towards unstructured interactions},
  author={Jiang, Hao and Wang, Zhanchi and Jin, Yusong and Chen, Xiaotong and Li, Peijin and Gan, Yinghao and Lin, Sen and Chen, Xiaoping},
  journal={The International Journal of Robotics Research},
  volume={40},
  number={1},
  pages={411--434},
  year={2021},
  publisher={SAGE Publications Sage UK: London, England}
}

@article{xie2023octopus,
  title={Octopus-inspired sensorized soft arm for environmental interaction},
  author={Xie, Zhexin and Yuan, Feiyang and Liu, Jiaqi and Tian, Lufeng and Chen, Bohan and Fu, Zhongqiang and Mao, Sizhe and Jin, Tongtong and Wang, Yun and He, Xia and others},
  journal={Science Robotics},
  volume={8},
  number={84},
  pages={eadh7852},
  year={2023},
  publisher={American Association for the Advancement of Science}
}

@article{nazeer2024rl,
  title={Rl-based adaptive controller for high precision reaching in a soft robot arm},
  author={Nazeer, Muhammad Sunny and Laschi, Cecilia and Falotico, Egidio},
  journal={IEEE Transactions on Robotics},
  volume={40},
  pages={2498--2512},
  year={2024},
  publisher={IEEE}
}

@article{li2024disturbance,
  title={Disturbance-adaptive tapered soft manipulator with precise motion controller for enhanced task performance},
  author={Li, Xianglong and Xiong, Quan and Sui, Dongbao and Zhang, Qinghua and Li, Hongwu and Wang, Ziqi and Zheng, Tianjiao and Wang, Hesheng and Zhao, Jie and Zhu, Yanhe},
  journal={IEEE Transactions on Robotics},
  volume={40},
  pages={3581--3601},
  year={2024},
  publisher={IEEE}
}

@article{wang2025robust,
  title={Robust Koopman-MPC approach with high-order disturbance observer for control of pneumatic soft bending actuators under external loads},
  author={Wang, Jiajin and Xu, Baoguo and Liu, Jinhao and Zhao, Zishuo and Peng, Weifeng and Song, Aiguo},
  journal={IEEE/ASME Transactions on Mechatronics},
  volume={30},
  number={6},
  pages={6455--6466},
  year={2025},
  publisher={IEEE}
}

@article{ho2018localized,
  title={Localized online learning-based control of a soft redundant manipulator under variable loading},
  author={Ho, Justin DL and Lee, Kit-Hang and Tang, Wai Lun and Hui, Ka-Ming and Althoefer, Kaspar and Lam, James and Kwok, Ka-Wai},
  journal={Advanced Robotics},
  volume={32},
  number={21},
  pages={1168--1183},
  year={2018},
  publisher={Taylor \& Francis}
}

@inproceedings{jitosho2023reinforcement,
  title={Reinforcement learning enables real-time planning and control of agile maneuvers for soft robot arms},
  author={Jitosho, Rianna and Lum, Tyler Ga Wei and Okamura, Allison and Liu, Karen},
  booktitle={Conference on Robot Learning},
  pages={1131--1153},
  year={2023},
  organization={PMLR}
}

@article{chen2024data,
  title={Data-driven methods applied to soft robot modeling and control: A review},
  author={Chen, Zixi and Renda, Federico and Le Gall, Alexia and Mocellin, Lorenzo and Bernabei, Matteo and Dangel, Th{\'e}o and Ciuti, Gastone and Cianchetti, Matteo and Stefanini, Cesare},
  journal={IEEE Transactions on Automation Science and Engineering},
  volume={22},
  pages={2241--2256},
  year={2024},
  publisher={IEEE}
}

@article{liu2025data,
  title={Data-driven methods for sensing, modeling and control of soft continuum robot: A review},
  author={Liu, Jiaqi and Duo, Youning and Chen, Xingyu and Zuo, Zonghao and Liu, Yuchen and Wen, Li},
  journal={IEEE/ASME Transactions on Mechatronics},
  year={2025},
  publisher={IEEE}
}

@article{liu2023underwater,
  title={An underwater robotic system with a soft continuum manipulator for autonomous aquatic grasping},
  author={Liu, Jiaqi and Song, Zhuheng and Lu, Yue and Yang, Hui and Chen, Xingyu and Duo, Youning and Chen, Bohan and Kong, Shihan and Shao, Zhuyin and Gong, Zheyuan and others},
  journal={IEEE/ASME Transactions on Mechatronics},
  volume={29},
  number={2},
  pages={1007--1018},
  year={2023},
  publisher={IEEE}
}

@article{huang2025grasping,
  title={Grasping by spiraling: reproducing elephant movements with rigid-soft robot synergy},
  author={Huang, Huishi and Wang, Haozhe and Fang, Chongyu and Yan, Mingge and Xu, Ruochen and Zhang, Yiyuan and Wang, Zhanchi and Ying, Fengkang and Liu, Jun and Laschi, Cecilia and others},
  journal={npj Robotics},
  volume={3},
  number={1},
  pages={18},
  year={2025},
  publisher={Nature Publishing Group UK London}
}

@inproceedings{huang2026physical,
  title={Physical Human-Robot Interaction for Grasping in Augmented Reality via Rigid-Soft Robot Synergy},
  author={Huang, Huishi and Klusmann, Jack and Wang, Haozhe and Ji, Shuchen and Ying, Fengkang and Zhang, Yiyuan and Nassour, John and Cheng, Gordon and Rus, Daniela and Liu, Jun and others},
  booktitle={2026 IEEE 9th International Conference on Soft Robotics (RoboSoft)},
  pages={1468--1475},
  year={2026},
  organization={IEEE}
}

@article{della2023model,
  title={Model-based control of soft robots: A survey of the state of the art and open challenges},
  author={Della Santina, Cosimo and Duriez, Christian and Rus, Daniela},
  journal={IEEE control systems magazine},
  volume={43},
  number={3},
  pages={30--65},
  year={2023},
  publisher={IEEE}
}

@article{liu2026underwater,
  title={Underwater soft arm grasping with simplified control using octopus-inspired bending propagation},
  author={Liu, Jiaqi and Zhu, Zhichao and Wen, Li},
  journal={npj Robotics},
  volume={4},
  number={1},
  pages={2},
  year={2026},
  publisher={Nature Publishing Group UK London}
}

@article{mazzolai2019octopus,
  title={Octopus-inspired soft arm with suction cups for enhanced grasping tasks in confined environments},
  author={Mazzolai, Barbara and Mondini, Alessio and Tramacere, Francesca and Riccomi, Gianluca and Sadeghi, Ali and Giordano, Goffredo and Del Dottore, Emanuela and Scaccia, Massimiliano and Zampato, Massimo and Carminati, Stefano},
  journal={Advanced Intelligent Systems},
  volume={1},
  number={6},
  pages={1900041},
  year={2019},
  publisher={Wiley Online Library}
}

@article{del2026peripheral,
  title={Peripheral control enabled by distributed sensing in an octopus-inspired soft robotic arm for autonomous underwater grasping},
  author={Del Dottore, Emanuela and Adhami, Romina and Shahabi, Ebrahim and Solfiti, Emanuele and Martini, Michele and Mariani, Stefano and Parmiggiani, Alberto and Mondini, Alessio and Sinibaldi, Edoardo and Mazzolai, Barbara},
  journal={Nature Machine Intelligence},
  pages={1--14},
  year={2026},
  publisher={Nature Publishing Group UK London}
}

@software{PyElastica,
  author       = {Arman Tekinalp and
                  Seung Hyun Kim and
                  Yashraj Bhosale and
                  Tejaswin Parthasarathy and
                  Noel Naughton and
                  Ali Albazroun and
                  Rahul Joon and
                  Songyuan Cui and
                  Ilia Nasiriziba and
                  Maximilian Stölzle and
                  Chia-Hsien (Cathy) Shih and
                  Mattia Gazzola},
  title        = {GazzolaLab/PyElastica},
  year         = 2024,
  publisher    = {Zenodo},
  doi          = {10.5281/zenodo.7658871},
  url          = {https://doi.org/10.5281/zenodo.7658871}
}

@article{zhang2026reinforcement,
  title={Reinforcement learning in linear embedding space unlocks generalizable control across soft robot configurations},
  author={Zhang, Xinglong and Li, Cong and Mo, Hangjie and Jiang, Yue and Cao, Wenyu and Xu, Xin and Jiang, Wei and Bing, Zhenshan and Yang, Yihe and Li, Xiaojian and others},
  journal={Nature Communications},
  year={2026},
  publisher={Nature Publishing Group}
}

@article{johnson2025zero,
  title={Zero-shot Whole-Body Manipulation with a Large-Scale Soft Robotic Torso via Guided Reinforcement Learning},
  author={Johnson, Curtis C and Alessi, Carlo and Falotico, Egidio and Killpack, Marc D},
  journal={arXiv preprint arXiv:2509.23556},
  year={2025}
}

@article{preti2023sensorized,
  title={Sensorized objects used to quantitatively study distal grasping in the African elephant},
  author={Preti, Matteo Lo and Beccai, Lucia},
  journal={Iscience},
  volume={26},
  number={9},
  year={2023},
  publisher={Elsevier}
}

@inproceedings{hou2026quantitative,
  title={A Quantitative Comparison of Centralised and Distributed Reinforcement Learning-Based Control for Soft Robotic Arms},
  author={Hou, Linxin and Wu, Qirui and Qin, Zhihang and Banerjee, Neil and Guo, Yongxin and Laschi, Cecilia},
  booktitle={2026 IEEE 9th International Conference on Soft Robotics (RoboSoft)},
  pages={475--480},
  year={2026},
  organization={IEEE}
}

@article{wang2025spirobs,
  title={SpiRobs: Logarithmic spiral-shaped robots for versatile grasping across scales},
  author={Wang, Zhanchi and Freris, Nikolaos M and Wei, Xi},
  journal={Device},
  volume={3},
  number={4},
  year={2025},
  publisher={Elsevier}
}

@article{montero2024mastering,
  title={Mastering Contact-rich Tasks by Combining Soft and Rigid Robotics with Imitation Learning},
  author={Montero, Mariano Ram{\'\i}rez and Shahabi, Ebrahim and Franzese, Giovanni and Kober, Jens and Mazzolai, Barbara and Della Santina, Cosimo},
  journal={arXiv preprint arXiv:2410.07787},
  year={2024}
}

@article{gan2022reinforcement,
  title={A reinforcement learning method for motion control with constraints on an HPN arm},
  author={Gan, Yinghao and Li, Peijin and Jiang, Hao and Wang, Gaotian and Jin, Yusong and Chen, Xiaoping and Ji, Jianmin},
  journal={IEEE Robotics and Automation Letters},
  volume={7},
  number={4},
  pages={12006--12013},
  year={2022},
  publisher={IEEE}
}

@article{tang2026general,
  title={A general soft robotic controller inspired by neuronal structural and plastic synapses that adapts to diverse arms, tasks, and perturbations},
  author={Tang, Zhiqiang and Tian, Liying and Xin, Wenci and Wang, Qianqian and Rus, Daniela and Laschi, Cecilia},
  journal={Science Advances},
  volume={12},
  number={2},
  pages={eaea3712},
  year={2026},
  publisher={American Association for the Advancement of Science}
}

@article{habich2025generalizable,
  title={Generalizable and fast surrogates: model predictive control of articulated soft robots using physics-informed neural networks},
  author={Habich, Tim-Lukas and Mohammad, Aran and Ehlers, Simon FG and Bensch, Martin and Seel, Thomas and Schappler, Moritz},
  journal={IEEE Transactions on Robotics},
  volume={42},
  pages={619--636},
  year={2025},
  publisher={IEEE}
}

@article{laschi2026soft,
  title={Soft robotics: what’s next in bioinspired design and applications of soft robots?},
  author={Laschi, Cecilia and Wen, Li and Iida, Fumiya and Abdulali, Arsen and Hauser, Helmut and Wang, Yifan and Liu, Ke and Ricotti, Leonardo and Cianchetti, Matteo and Althoefer, Kaspar and others},
  journal={Bioinspiration \& Biomimetics},
  volume={21},
  number={1},
  pages={011501},
  year={2026},
  publisher={IOP Publishing}
}

@article{tang2026learning,
  title={Learning to control a soft robotic manipulator under uncertainty and unforeseen changes in robot--environment interaction},
  author={Tang, Zhiqiang and Wang, Peiyi and Xin, Wenci and Laschi, Cecilia},
  journal={The International Journal of Robotics Research},
  volume={45},
  number={3},
  pages={452--476},
  year={2026},
  publisher={SAGE Publications Sage UK: London, England}
}

@article{alessi2024pushing,
  title={Pushing with soft robotic arms via deep reinforcement learning},
  author={Alessi, Carlo and Bianchi, Diego and Stano, Gianni and Cianchetti, Matteo and Falotico, Egidio},
  journal={Advanced Intelligent Systems},
  volume={6},
  number={8},
  pages={2300899},
  year={2024},
  publisher={Wiley Online Library}
}

@article{buresch2022contact,
  title={Contact chemoreception in multi-modal sensing of prey by Octopus},
  author={Buresch, KC and Sklar, K and Chen, JY and Madden, SR and Mongil, AS and Wise, GV and Boal, JG and Hanlon, RT},
  journal={Journal of Comparative Physiology A},
  volume={208},
  number={3},
  pages={435--442},
  year={2022},
  publisher={Springer}
}

@article{fischer2023dynamic,
  title={Dynamic Task Space Control Enables Soft Manipulators to Perform Real-World Tasks},
  author={Fischer, Oliver and Toshimitsu, Yasunori and Kazemipour, Amirhossein and Katzschmann, Robert K},
  journal={Advanced Intelligent Systems},
  volume={5},
  number={1},
  pages={2200024},
  year={2023},
  publisher={Wiley Online Library}
}

@inproceedings{satheeshbabu2019open,
  title={Open loop position control of soft continuum arm using deep reinforcement learning},
  author={Satheeshbabu, Sreeshankar and Uppalapati, Naveen Kumar and Chowdhary, Girish and Krishnan, Girish},
  booktitle={2019 International Conference on Robotics and Automation (ICRA)},
  pages={5133--5139},
  year={2019},
  organization={IEEE}
}

@article{centurelli2022closed,
  title={Closed-loop dynamic control of a soft manipulator using deep reinforcement learning},
  author={Centurelli, Andrea and Arleo, Luca and Rizzo, Alessandro and Tolu, Silvia and Laschi, Cecilia and Falotico, Egidio},
  journal={IEEE Robotics and Automation Letters},
  volume={7},
  number={2},
  pages={4741--4748},
  year={2022},
  publisher={IEEE}
}

@article{guan2023trimmed,
  title={Trimmed helicoids: an architectured soft structure yielding soft robots with high precision, large workspace, and compliant interactions},
  author={Guan, Qinghua and Stella, Francesco and Della Santina, Cosimo and Leng, Jinsong and Hughes, Josie},
  journal={npj Robotics},
  volume={1},
  number={1},
  pages={4},
  year={2023},
  publisher={Nature Publishing Group UK London}
}

@article{rus2015design,
  title={Design, fabrication and control of soft robots},
  author={Rus, Daniela and Tolley, Michael T},
  journal={Nature},
  volume={521},
  number={7553},
  pages={467--475},
  year={2015},
  publisher={Nature Publishing Group UK London}
}

@article{trivedi2008soft,
  title={Soft robotics: Biological inspiration, state of the art, and future research},
  author={Trivedi, Deepak and Rahn, Christopher D and Kier, William M and Walker, Ian D},
  journal={Applied bionics and biomechanics},
  volume={5},
  number={3},
  pages={99--117},
  year={2008},
  publisher={Taylor \& Francis}
}

@article{bruder2025koopman,
  title={A Koopman-based residual modeling approach for the control of a soft robot arm},
  author={Bruder, Daniel and Bombara, David and Wood, Robert J},
  journal={The International journal of robotics research},
  volume={44},
  number={3},
  pages={388--406},
  year={2025},
  publisher={Sage Publications Sage UK: London, England}
}

@article{thuruthel2018model,
  title={Model-based reinforcement learning for closed-loop dynamic control of soft robotic manipulators},
  author={Thuruthel, Thomas George and Falotico, Egidio and Renda, Federico and Laschi, Cecilia},
  journal={IEEE Transactions on Robotics},
  volume={35},
  number={1},
  pages={124--134},
  year={2018},
  publisher={IEEE}
}

@article{franco2021energy,
  title={Energy-shaping control of soft continuum manipulators with in-plane disturbances},
  author={Franco, Enrico and Garriga-Casanovas, Arnau},
  journal={The International Journal of Robotics Research},
  volume={40},
  number={1},
  pages={236--255},
  year={2021},
  publisher={Sage Publications Sage UK: London, England}
}

@article{whitesides2018soft,
author = {Whitesides, George M.},
title = {Soft Robotics},
journal = {Angewandte Chemie International Edition},
volume = {57},
number = {16},
pages = {4258-4273},
year = {2018}
}

@article{laschi2016soft,
  title={Soft robotics: Technologies and systems pushing the boundaries of robot abilities},
  author={Laschi, Cecilia and Mazzolai, Barbara and Cianchetti, Matteo},
  journal={Science robotics},
  volume={1},
  number={1},
  pages={eaah3690},
  year={2016},
  publisher={American Association for the Advancement of Science}
}

@article{wang2024sensing,
  title={Sensing expectation enables simultaneous proprioception and contact detection in an intelligent soft continuum robot},
  author={Wang, Peiyi and Xie, Zhexin and Xin, Wenci and Tang, Zhiqiang and Yang, Xinhua and Mohanakrishnan, Muralidharan and Guo, Sheng and Laschi, Cecilia},
  journal={Nature Communications},
  volume={15},
  number={1},
  pages={9978},
  year={2024},
  publisher={Nature Publishing Group UK London}
}

@article{bang2024bioinspired,
  title={Bioinspired electronics for intelligent soft robots},
  author={Bang, Junhyuk and Choi, Seok Hwan and Pyun, Kyung Rok and Jung, Yeongju and Hong, Sangwoo and Kim, Dohyung and Lee, Youngseok and Won, Daeyeon and Jeong, Seongmin and Shin, Wooseop and others},
  journal={Nature Reviews Electrical Engineering},
  volume={1},
  number={9},
  pages={597--613},
  year={2024},
  publisher={Nature Publishing Group UK London}
}
\bibliographystyle{sciencemag}

\end{document}